\documentclass[12pt]{amsart}
\usepackage{lmodern} %czcionka
\usepackage[utf8]{inputenc} 
\usepackage[T1]{fontenc}
\usepackage{amsfonts}
\usepackage{amsmath}
\usepackage{amssymb}
\usepackage{amsthm}
\usepackage[a4paper, left=3cm, right=3cm, top=3cm,
bottom=3cm,dvips]{geometry}
\usepackage{graphicx}
\usepackage[font=small,labelfont=bf]{caption}
\usepackage{color}
\usepackage[normalem]{ulem}
\usepackage{textcomp}
\usepackage{xcolor}
\usepackage{multirow}

\title[End-to-End Convolutional Activation Anomaly Analysis]{End-to-End Convolutional Activation Anomaly Analysis for Anomaly Detection}
\author[A. Kozłowski, D. Ponikowski, P. Żukiewicz, P. Twardowski]{Aleksander Kozłowski, Daniel Ponikowski, Piotr Żukiewicz, Paweł Twardowski*\\ \\Artificial Intelligence Team\\Samsung R\&D Institute Poland}

\begin{document}
	
\begin{abstract}
    We propose an End-to-end Convolutional Activation Anomaly Analysis (E2E-CA$^3$), which is a significant extension of A$^3$ anomaly detection approach proposed by Sperl, Schulze and B{\"o}ttinger~\cite{A3}, both in terms of architecture and scope of application. In contrast to the original idea, we utilize a convolutional autoencoder as a target network, which allows for natural application of the method both to image and tabular data. The alarm network is also designed as a CNN, where the activations of convolutional layers from CAE are stacked together into $k+1-$dimensional tensor. Moreover, we combine the classification loss of the alarm network with the reconstruction error of the target CAE, as a "best of both worlds" approach, which greatly increases the versatility of the network. The evaluation shows that despite generally straightforward and lightweight architecture, it has a very promising anomaly detection performance on common datasets such as MNIST, CIFAR-10 and KDDcup99. 
\end{abstract}

\vspace{150pt}

\maketitle

\section{Introduction}
The problem of anomaly detection (AD) regards identifying observations that are abnormal or outlying in the context of the ones regarded "normal" or "typical" to the dataset. It is of great interest in many real-life areas and scenarios, ranging from fraud/intrusion detection to medical imagining and industrial data quality control. This results in a large effort in designing accurate and efficient anomaly detection algorithms.

Machine learning (ML) and deep learning (DL) techniques have consistently shown strong results in anomaly detection tasks. Autoencoders (AE) are the most widespread type of deep neural networks (DNN) for the anomaly detection purposes. While trained only on typical data, they learn a relatively compact, low-dimensional representation of the data, retaining mostly the information that is fairly common among the observations. Conversely, it will be unable to grasp atypical features of the anomalous samples, leading to faulty reconstruction, hence large reconstruction error. However, for this approach to be successful, a clear boundary between reconstruction errors of typical and anomalous data is necessary, which is not always the case, as the information about, possibly very high-dimensional, data is restricted to a single value. As a possible alleviation to these limitations, Sperl, Schulze and B{\"o}ttinger proposed the concept of Activation Anomaly Analysis (A$^3$)~\cite{A3}. Intuitively, their approach is based on the assumption that the activations of a neural network (for example an autoencoder) trained only on typical data will exhibit unusual patterns when the network is presented with an anomalous example. This network is called a \emph{target network} Therefore, an auxilliary network, called \emph{alarm network}, can be constructed and trained using the activations as the inputs. The binary output from the auxilliary network acts as a flag for potentially outlying observations. The approach is noteworthy for its impressive complexity-to-performance ratio, as from an architectural standpoint is just a feed-forward network.

The drawback of this approach is that it omits the information conveyed by the reconstruction error of the underlying target autoencoder. Even if that information may be in general weaker that the one inferred from the behaviour of the activation, it can still serve as a valuable supplement. We there propose an amalgamation of the classic autoencoder-based anomaly detection approach with the activation analysis techniques to construct an efficient method of supervised anomaly detection. On top of that, we modify the costruction presented in~\cite{A3} by utilizing the convolutional architectures, which extends the scope of application of the models beyond tabular data also to images. Moreover, the CNN approach furthermore decreases the number of parameters compared to the original approach with parameter sharing. We compare the results of our method with the original A$^3$ approach as well as with other state-of-the-art methods of similar complexity, with very promising outcomes.

\section{Related work}

As anomaly detection has major applications in science, medicine, industry etc., a variety of techniques and methods were developed over the years. This includes basic statistical methods such as Grubbs' test, more advanced ML-based algorithms such as Isolation Forests, Local Outlier Factor, Elliptic Envelopes and One-class SVMs, as well as NN and DL models, which are currently a main focus of a lively and widespread research. 

Utilization of autoencoders is the most popular direction in NN- and DL-based approach to AD. Autoencoders can be used in this context as feature extractors~\cite{FAE1,FAE2,FAE3}, but commonly the reconstruction error is taken as an indicator of anomalous observation, with high-error instances labelled as anomalies. The rationale for this is that anomalies lack the common features that are highly compressed in the latent space, therefore their reconstruction is less precise than that of regular observations~\cite{AE1,AE2,AE3,AE4}. Variational autoencoders (VAE) prove to be especially interesting in this context~\cite{VAE1,VAE2,VAE3,VAE4}. Application of RNN-based autoencoders is also considered~\cite{RAE}. Other NN/DL approaches to the problem can be found for example in~\cite{devnet,mkd}, and~\cite{gan}, while \cite{anogan} and~\cite{fanogan} showcase the utilization of generative models in order to achieve a good performance in an unsupervised setting. 

The drawback of the reconstruction loss autoencoder approach is that it is highly dependent on the choice of the loss threshold, the breach of which indicates an anomaly. To counter this problem, Sperl, Schulze and B{\"o}ttinger introduced the A$^3$ method described earlier. They then extended this approach to the generative setting in~\cite{gana3}, where they also utilized convolutional autoencoders to some extent. The also suggested a similar method that utilizes gradients instead of activations~\cite{r2d2,da3g}, and a method utilizing unsupervised experts~\cite{gated}. We present an alternative extension of their work, introducing the CNNs in broader manner and combining the activation analysis with reconstruction loss. For surveys on the topic, see \cite{survey1, survey2}.

\section{Methodology}

\subsection{Architecture}

\begin{figure}
	\centering
	\includegraphics[width=0.5\textwidth]{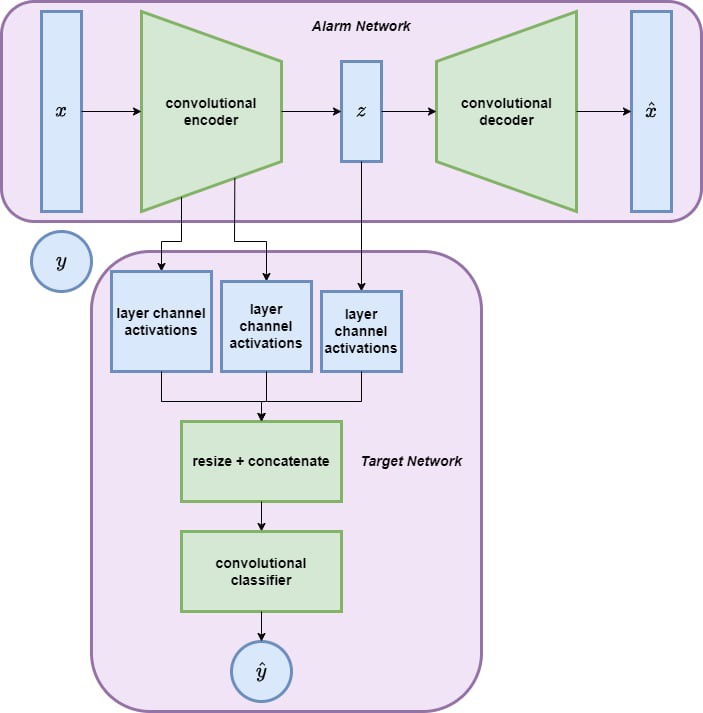}
	\caption{Architecture diagram}
	\label{fig:arch_schema}
\end{figure}

Figure~\ref{fig:arch_schema} presents the architecture of our proposed solution. We follow the lead set up in the original paper: we assume the hypothesis that the activations of the model trained solely on the data without anomalies will showcase distinguishable patterns when presented with an anomalous example - one which was not seen during the training phase. The patterns can be then understood by a secondary binary model trained to separate the observations based on the activations. To achieve this, we construct a network that comprises of two convolutional subnetworks. The \emph{target network}, trained only on typical examples, is a Convolutional Variational Autoencoder that realizes the primary objective of encoding the observations onto the latent space and then reconstructing them to the original format with the goal of minimizing the reconstruction loss. The \emph{alarm network} is a standard feed-forward Convolutional Neural Network with the binary classifier output. The input to this network is a concatenation (in the channel dimension) of the activation tensors from the individual layers and channels.

This concatenation is not a trivial issue and can be envisioned in several ways. The main difficulty is that tensors with activations from each convolutional layer can differ in shape and thus cannot be concatenated right away. Therefore a form of reshaping such as flattening, padding or pooling, is needed. In our case pooling the activations to the minimal size in the set of concatenated tensors proved to be the optimal solution. This preserves channel-like structure of underlying data and doesn't introduce redundant parameters which would appear in the case of padding.

\subsection{Convolutional layers}
It is obvious that image data differs from tabular. The latter one is unstructured and observations are represented as vectors of features, whereas images are structured (have defined geometry) and are encoded as a 3-dimensional tensors. As one of the main assumptions for our model is to utilize convolutional architecture regardless of the nature of the input, a split approach was set up.

\subsubsection{Image data}

In the case of image data we simply use 2-dimensional convolution layers in the encoder and 2-dimensional deconvolution layers in the decoder. As a result, target network operates with 3-dimensional convolution layers.

\subsubsection{Tabular data}

In order for the convolution usage to make sense, the data has to have some structure. To introduce it in the case of tabular data, we use the trick called \emph{soft-ordering} \cite{softordering}. We take the input vector and instead of passing it directly to encoder, we firstly pass it through a single linear layer with \emph{ELU} activation function. In the learning process, this layer learns a transformation that creates a structured vector in a way that is suitable for the task. We subsequently split the vector in $k$ parts and stack them on top of each other. The result is a 2-dimensional tensor that can be passed through a 1-dimensional convolution layers, which are the building blocks for encoder and decoder as well as target network.

Introducing such operations to the encoder forces us to also learn their "reverse", thus reshaping and linear layer was added before the output layer of the decoder. 

\subsection{End-to-end learning}

In the standard A$^3$ approach the target and alarm networks have disjoint learning phases - the target network is trained as the first step, and the alarm network is trained subsequently. It turned out, however, that it is beneficial and more convenient to allow the information to flow between the networks during the learning stage. Consequently simultaneous, end-to-end learning process was designed.

In order to train both alarm and target networks at the same time we simply combine their losses. The model aims to minimize following expression:
\begin{equation}
	\mathcal{L}(x, y, \hat{x}, \hat{y}) = \mathcal{L}_{recon}(x, \hat{x}) + \gamma \cdot \mathcal{L}_{clf}(y, \hat{y})
\end{equation}
where $\gamma>0$ is a hyperparameter. That hyperparameter defines the amount of information allowed to flow through one of the components relative to the other. Taking most commonly used reconstruction error and classification losses leaves us with:
\begin{equation}
	\mathcal{L}(x, y, \hat{x}, \hat{y}) = \text{MSE}(x, \hat{x}) + \gamma \cdot \text{BCE}(y, \hat{y}).
\end{equation}

To make sure we align with the assumptions of $A^3$ framework by not feeding the autoencoder with anomalous samples, we put the reconstruction loss as 0 for these observations. This way we give the encoder the power to readjust its weights with regard to the classification task. 

\section{Experiments}

\subsection{Setup}

\subsubsection{Image datasets}\hfill\\

\textbf{MNIST}~\cite{mnist} is a popular benchmark dataset with 70,000 grayscale 28x28 images of handwritten digits, split between training and test sets in a 6:1 ratio. It is typically used in multiclass classification problems, therefore, to make it suitable for an anomaly detection task, the 'one-vs-all' approach was employed. The representatives of each of the 10 classes were successively treated as typical examples, while the remaining nine digits served as anomalies. Separate instances of the end-to-end network were then trained for each set of anomalous digits.

\textbf{CIFAR-10}~\cite{cifar10} contains 50,000 training and 10,000 test images (three channels, 32x32 pixels), divided equally between 10 classes of objects (e.g. airplanes, ships, cats, dogs). The dataset was used to formulate a one-vs-all anomaly detection problem.

Similarly to MNIST, \textbf{Fashion-MNIST}~\cite{fmnist} is a set of 28x28 grayscale images with 60,000 training and 10,000 test examples. Images from this dataset belong to one of 10 classes representing pieces of clothing (e.g. shirt, coat, dress, etc.). Once again, the anomaly detection task was performed separately with each class acting as a typical one, while the remaining classes were used as examples of anomalous objects.

\subsubsection{Tabular datasets}\hfill\\

\textbf{Credit Card Fraud Detection}~\cite{ccfraud} dataset contains transactions made by credit cards within two days in September 2013. Total number of transactions is 284,807 from which only 492, or in other words $0.17\%$, were fraudulent. We consider all of the 28 PCA-encoded features as well as those named 'Time' and 'Amount'.

\textbf{KDDcup99}~\cite{kdd} was used for The Third International Knowledge Discovery and Data Mining Tools Competition in which the task was to separate normal connections from intrusions. The entire dataset contains 4,898,431 observations, but for our purposes we derive two versions which we call \textit{kdd(http)} and \textit{kdd(ps)} respectively. The first one is created by filtering the whole dataset to keep only \textit{http} connections, whereas to obtain the second one, we only keep \textit{portsweep} as the intrusion type. After dropping duplicates we are left with 580,507 and 816,378 observations respectively, from which $0.48\%$ and $0.44\%$ are labeled as anomalies (attacks).

\textbf{CelebA}~\cite{liu2015faceattributes} dataset contains images of more than 200,000 celebrities. Rather than representing image as pixels, each sample is encoded as 40 binary annotated attributes. We consider the \textit{bald} attribute to be our target, by doing so we are left with observations of 39 features, from which $3.12\%$ are considered anomalous.

\textbf{Census-Income (KDD)}~\cite{misc_census-income_(kdd)_117} contains census data gathered by the U.S. Census Bureau in 1994-1995. As anomalous observations we consider people with high annual earnings (more than 50K US dollars). Dataset contains 299,285 observation and anomaly class is about $8.30\%$. 

In \textbf{Bank Marketing}~\cite{misc_bank_marketing_222} dataset we find data from Portuguese banking institution about direct marketing campaign based on phone calls. Records with successful campaigns are considered as anomaly class ($11.27\%$). Entire dataset contains 45,211 observations.

\textbf{Thyroid Disease}~\cite{misc_thyroid_disease_102} dataset is a collection of 10 separate databases from Garvan Institute in Sydney, Australia. Dataset contains 7,200 observations among which 534 ($7.49\%$) are patients diagnosed with hypothyroid. These patients were labelled as anomalies.

To be consistent across all datasets, even if the dataset had separate test set we didn't use it in out experiments. Each dataset (or a train part if test was available) was split between 3 parts: train, validation and test, with a 8:1:1 ratio. In order to maintain relationship between the number of typical and anomalous samples in each split, we used stratified sampling technique.

\begin{table*}[t]
	\centering
	\caption{Table with results for E2E-ConvA$^3$ and other compared methods obtained on image datasets.}
	\label{tab:image_results}
	\resizebox{\textwidth}{!}{%
		\begin{tabular}{|cl|llllll|}
			\hline
			\multicolumn{2}{|c|}{} &
			\multicolumn{6}{c|}{AUC-ROC perofrmance on test datasets} \\ \hline
			\multicolumn{1}{|l|}{Data} &
			Normal class &
			%\multicolumn{1}{l|}{CNN classifier} &
			\multicolumn{1}{l|}{Isolation Forest\footnote{for Isolation Forest we use following formula: $\max(x, 1-x)$, where $x$ means AUR-ROC performance on test dataset}} &
			\multicolumn{1}{l|}{MKD-AD~\cite{mkd}} &
			\multicolumn{1}{l|}{DSVDD~\cite{DSVDD,mkd}} &
			\multicolumn{1}{l|}{DASVDD~\cite{DASVDD}} &
			\multicolumn{1}{l|}{E2E-ConvA$^3$ ($\gamma=1$)} &
			E2E-ConvA$^3$ ($\gamma=10$) \\ \hline \hline
			\multicolumn{1}{|c|}{} &
			0 &
			%\multicolumn{1}{l|}{0.9990} &
			\multicolumn{1}{l|}{0.8141} &
			\multicolumn{1}{l|}{0.9982} &
			\multicolumn{1}{l|}{0.980} &
			\multicolumn{1}{l|}{0.997} &
			\multicolumn{1}{l|}{0.9993 ±   0.0008} &
			\textbf{0.9994 ± 0.0007} \\ \cline{2-8} 
			\multicolumn{1}{|c|}{} &
			1 &
			%\multicolumn{1}{l|}{0.9999} &
			\multicolumn{1}{l|}{0.9598} &
			\multicolumn{1}{l|}{0.9982} &
			\multicolumn{1}{l|}{0.997} &
			\multicolumn{1}{l|}{0.999} &
			\multicolumn{1}{l|}{0.9996 ± 0.0005} &
			\textbf{0.9998 ± 0.0003} \\ \cline{2-8}
			\multicolumn{1}{|c|}{} &
			2 &
			%\multicolumn{1}{l|}{0.9995} &
			\multicolumn{1}{l|}{0.7148} &
			\multicolumn{1}{l|}{0.9779} &
			\multicolumn{1}{l|}{0.917} &
			\multicolumn{1}{l|}{0.954} &
			\multicolumn{1}{l|}{0.9986 ± 0.0008} &
			\textbf{0.9988 ± 0.0009} \\ \cline{2-8}
			\multicolumn{1}{|c|}{} &
			3 &
			%\multicolumn{1}{l|}{\textbf{1.0000}} &
			\multicolumn{1}{l|}{0.5866} &
			\multicolumn{1}{l|}{0.9875} &
			\multicolumn{1}{l|}{0.919} &
			\multicolumn{1}{l|}{0.962} &
			\multicolumn{1}{l|}{0.9979 ± 0.0008} &
			\textbf{0.9984 ± 0.0012} \\ \cline{2-8}
			\multicolumn{1}{|c|}{} &
			4 &
			%\multicolumn{1}{l|}{\textbf{1.0000}} &
			\multicolumn{1}{l|}{0.5673} &
			\multicolumn{1}{l|}{0.9843} &
			\multicolumn{1}{l|}{0.949} &
			\multicolumn{1}{l|}{0.981} &
			\multicolumn{1}{l|}{0.9982 ± 0.0018} &
			\textbf{0.9989 ± 0.0009} \\ \cline{2-8}
			\multicolumn{1}{|c|}{} &
			5 &
			%\multicolumn{1}{l|}{0.9988} &
			\multicolumn{1}{l|}{0.5700} &
			\multicolumn{1}{l|}{0.9816} &
			\multicolumn{1}{l|}{0.885} &
			\multicolumn{1}{l|}{0.972} &
			\multicolumn{1}{l|}{0.9984 ± 0.0011} &
			\textbf{0.9988 ± 0.0011} \\ \cline{2-8}
			\multicolumn{1}{|c|}{} &
			6 &
			%\multicolumn{1}{l|}{0.9978} &
			\multicolumn{1}{l|}{0.6080} &
			\multicolumn{1}{l|}{0.9943} &
			\multicolumn{1}{l|}{0.983} &
			\multicolumn{1}{l|}{0.996} &
			\multicolumn{1}{l|}{\textbf{0.9989 ± 0.0007}} &
			\textbf{0.9989 ± 0.0010} \\ \cline{2-8}
			\multicolumn{1}{|c|}{} &
			7 &
			%\multicolumn{1}{l|}{0.9990} &
			\multicolumn{1}{l|}{0.6099} &
			\multicolumn{1}{l|}{0.9838} &
			\multicolumn{1}{l|}{0.946} &
			\multicolumn{1}{l|}{0.981} &
			\multicolumn{1}{l|}{0.9985 ± 0.0008} &
			\textbf{0.9986 ± 0.0007} \\ \cline{2-8}
			\multicolumn{1}{|c|}{} &
			8 &
			%\multicolumn{1}{l|}{0.9989} &
			\multicolumn{1}{l|}{0.5430} &
			\multicolumn{1}{l|}{0.9841} &
			\multicolumn{1}{l|}{0.939} &
			\multicolumn{1}{l|}{0.942} &
			\multicolumn{1}{l|}{0.9985 ± 0.0010} &
			\textbf{0.9989 ± 0.0009} \\ \cline{2-8}
			\multicolumn{1}{|c|}{} &
			9 &
			%\multicolumn{1}{l|}{0.9970} &
			\multicolumn{1}{l|}{0.6505} &
			\multicolumn{1}{l|}{0.9810} &
			\multicolumn{1}{l|}{0.965} &
			\multicolumn{1}{l|}{0.983} &
			\multicolumn{1}{l|}{0.9972 ± 0.0021} &
			\textbf{0.9973 ± 0.0014} \\ \cline{2-8}
			\multicolumn{1}{|c|}{\multirow{-11}{*}{MNIST}} &
			mean &
			%\multicolumn{1}{l|}{0.9989} &
			\multicolumn{1}{l|}{0.6624} &
			\multicolumn{1}{l|}{0.9870} &
			\multicolumn{1}{l|}{0.9479} &
			\multicolumn{1}{l|}{0.977} &
			\multicolumn{1}{l|}{0.9982} &
			\textbf{0.9986} \\ \hline \hline
			\multicolumn{1}{|c|}{} &
			0 &
			%\multicolumn{1}{l|}{\textbf{0.9893}} &
			\multicolumn{1}{l|}{0.5696} &
			\multicolumn{1}{l|}{0.9250} &
			\multicolumn{1}{l|}{0.982} &
			\multicolumn{1}{l|}{0.912} &
			\multicolumn{1}{l|}{0.9847 ± 0.0035} &
			\textbf{0.9849 ± 0.0027} \\ \cline{2-8}
			\multicolumn{1}{|c|}{} &
			1 &
			%\multicolumn{1}{l|}{\textbf{0.9999}} &
			\multicolumn{1}{l|}{0.7839} &
			\multicolumn{1}{l|}{0.9921} &
			\multicolumn{1}{l|}{0.903} &
			\multicolumn{1}{l|}{0.990} &
			\multicolumn{1}{l|}{\textbf{0.9993 ± 0.0005}} &
			0.9986 ± 0.0009 \\ \cline{2-8}
			\multicolumn{1}{|c|}{} &
			2 &
			%\multicolumn{1}{l|}{\textbf{0.9875}} &
			\multicolumn{1}{l|}{0.6418} &
			\multicolumn{1}{l|}{0.9248} &
			\multicolumn{1}{l|}{0.907} &
			\multicolumn{1}{l|}{0.893} &
			\multicolumn{1}{l|}{\textbf{0.9869 ± 0.0036}} &
			0.9863 ± 0.0022 \\ \cline{2-8}
			\multicolumn{1}{|c|}{} &
			3 &
			%\multicolumn{1}{l|}{0.9937} &
			\multicolumn{1}{l|}{0.7237} &
			\multicolumn{1}{l|}{0.9380} &
			\multicolumn{1}{l|}{0.942} &
			\multicolumn{1}{l|}{0.937} &
			\multicolumn{1}{l|}{0.9915 ± 0.0035} &
			\textbf{0.9942 ± 0.0016} \\ \cline{2-8}
			\multicolumn{1}{|c|}{} &
			4 &
			%\multicolumn{1}{l|}{0.9865} &
			\multicolumn{1}{l|}{0.5110} &
			\multicolumn{1}{l|}{0.9295} &
			\multicolumn{1}{l|}{0.894} &
			\multicolumn{1}{l|}{0.907} &
			\multicolumn{1}{l|}{\textbf{0.9873 ± 0.0025}} &
			0.9855 ± 0.0033 \\ \cline{2-8}
			\multicolumn{1}{|c|}{} &
			5 &
			%\multicolumn{1}{l|}{0.9977} &
			\multicolumn{1}{l|}{0.5093} &
			\multicolumn{1}{l|}{0.9821} &
			\multicolumn{1}{l|}{0.918} &
			\multicolumn{1}{l|}{0.938} &
			\multicolumn{1}{l|}{0.9981 ± 0.0017} &
			\textbf{0.9982 ± 0.0011} \\ \cline{2-8}
			\multicolumn{1}{|c|}{} &
			6 &
			%\multicolumn{1}{l|}{\textbf{0.9720}} &
			\multicolumn{1}{l|}{0.5337} &
			\multicolumn{1}{l|}{0.8487} &
			\multicolumn{1}{l|}{0.834} &
			\multicolumn{1}{l|}{0.828} &
			\multicolumn{1}{l|}{0.9642 ± 0.0053} &
			\textbf{0.9662 ± 0.0050} \\ \cline{2-8}
			\multicolumn{1}{|c|}{} &
			7 &
			%\multicolumn{1}{l|}{0.9971} &
			\multicolumn{1}{l|}{0.7597} &
			\multicolumn{1}{l|}{0.9902} &
			\multicolumn{1}{l|}{0.988} &
			\multicolumn{1}{l|}{0.986} &
			\multicolumn{1}{l|}{0.9982 ± 0.0008} &
			\textbf{0.9983 ± 0.0007} \\ \cline{2-8}
			\multicolumn{1}{|c|}{} &
			8 &
			%\multicolumn{1}{l|}{0.9981} &
			\multicolumn{1}{l|}{0.8736} &
			\multicolumn{1}{l|}{0.9433} &
			\multicolumn{1}{l|}{0.919} &
			\multicolumn{1}{l|}{0.894} &
			\multicolumn{1}{l|}{0.9982 ± 0.0014} &
			\textbf{0.9986 ± 0.0013} \\ \cline{2-8}
			\multicolumn{1}{|c|}{} &
			9 &
			%\multicolumn{1}{l|}{\textbf{0.9987}} &
			\multicolumn{1}{l|}{0.6931} &
			\multicolumn{1}{l|}{0.9751} &
			\multicolumn{1}{l|}{0.990} &
			\multicolumn{1}{l|}{0.979} &
			\multicolumn{1}{l|}{0.9982 ± 0.0011} &
			\textbf{0.9988 ± 0.0003} \\ \cline{2-8}
			\multicolumn{1}{|c|}{\multirow{-11}{*}{Fashion-MNIST}} &
			mean &
			%\multicolumn{1}{l|}{\textbf{0.9921}} &
			\multicolumn{1}{l|}{0.6599} &
			\multicolumn{1}{l|}{0.9448} &
			\multicolumn{1}{l|}{0.9277} &
			\multicolumn{1}{l|}{0.926} &
			\multicolumn{1}{l|}{0.9906} &
			\textbf{0.9909} \\ \hline \hline
			\multicolumn{1}{|c|}{} &
			airplane &
			%\multicolumn{1}{l|}{\textbf{0.9567}} &
			\multicolumn{1}{l|}{0.5292} &
			\multicolumn{1}{l|}{0.9053} &
			\multicolumn{1}{l|}{0.617} &
			\multicolumn{1}{l|}{0.686} &
			\multicolumn{1}{l|}{0.9297 ± 0.0088} &
			\textbf{0.9322 ± 0.0099} \\ \cline{2-8}
			\multicolumn{1}{|c|}{} &
			automobile &
			%\multicolumn{1}{l|}{\textbf{0.9747}} &
			\multicolumn{1}{l|}{0.6412} &
			\multicolumn{1}{l|}{0.9035} &
			\multicolumn{1}{l|}{0.659} &
			\multicolumn{1}{l|}{0.643} &
			\multicolumn{1}{l|}{0.9643 ± 0.0059} &
			\textbf{0.9657 ± 0.0063} \\ \cline{2-8}
			\multicolumn{1}{|c|}{} &
			bird &
			%\multicolumn{1}{l|}{\textbf{0.9055}} &
			\multicolumn{1}{l|}{0.6229} &
			\multicolumn{1}{l|}{0.7966} &
			\multicolumn{1}{l|}{0.508} &
			\multicolumn{1}{l|}{0.558} &
			\multicolumn{1}{l|}{\textbf{0.8779 ± 0.0075}} &
			0.8771 ± 0.0103 \\ \cline{2-8}
			\multicolumn{1}{|c|}{} &
			cat &
			%\multicolumn{1}{l|}{\textbf{0.8980}} &
			\multicolumn{1}{l|}{0.5751} &
			\multicolumn{1}{l|}{0.7702} &
			\multicolumn{1}{l|}{0.591} &
			\multicolumn{1}{l|}{0.586} &
			\multicolumn{1}{l|}{0.8412 ± 0.0072} &
			\textbf{0.8495 ± 0.0118} \\ \cline{2-8}
			\multicolumn{1}{|c|}{} &
			deer &
			%\multicolumn{1}{l|}{\textbf{0.9427}} &
			\multicolumn{1}{l|}{0.6909} &
			\multicolumn{1}{l|}{0.8671} &
			\multicolumn{1}{l|}{0.609} &
			\multicolumn{1}{l|}{0.640} &
			\multicolumn{1}{l|}{0.9035 ± 0.0131} &
			\textbf{0.9074 ± 0.0064} \\ \cline{2-8}
			\multicolumn{1}{|c|}{} &
			dog &
			%\multicolumn{1}{l|}{\textbf{0.9223}} &
			\multicolumn{1}{l|}{0.5609} &
			\multicolumn{1}{l|}{\textbf{0.9140}} &
			\multicolumn{1}{l|}{0.657} &
			\multicolumn{1}{l|}{0.626} &
			\multicolumn{1}{l|}{0.8862 ± 0.0101} &
			0.8882 ± 0.0133 \\ \cline{2-8}
			\multicolumn{1}{|c|}{} &
			frog &
			%\multicolumn{1}{l|}{\textbf{0.9629}} &
			\multicolumn{1}{l|}{0.6351} &
			\multicolumn{1}{l|}{0.8898} &
			\multicolumn{1}{l|}{0.677} &
			\multicolumn{1}{l|}{0.710} &
			\multicolumn{1}{l|}{\textbf{0.9339 ± 0.0096}} &
			0.9327 ± 0.0120 \\ \cline{2-8}
			\multicolumn{1}{|c|}{} &
			horse &
			%\multicolumn{1}{l|}{\textbf{0.9517}} &
			\multicolumn{1}{l|}{0.5072} &
			\multicolumn{1}{l|}{0.8678} &
			\multicolumn{1}{l|}{0.673} &
			\multicolumn{1}{l|}{0.646} &
			\multicolumn{1}{l|}{0.9316 ± 0.0083} &
			\textbf{0.9318 ± 0.0099} \\ \cline{2-8}
			\multicolumn{1}{|c|}{} &
			ship &
			%\multicolumn{1}{l|}{\textbf{0.9727}} &
			\multicolumn{1}{l|}{0.5177} &
			\multicolumn{1}{l|}{0.9145} &
			\multicolumn{1}{l|}{0.759} &
			\multicolumn{1}{l|}{0.811} &
			\multicolumn{1}{l|}{0.9645 ± 0.0065} &
			\textbf{0.9646 ± 0.0083} \\ \cline{2-8} 
			\multicolumn{1}{|c|}{} &
			truck &
			%\multicolumn{1}{l|}{\textbf{0.9730}} &
			\multicolumn{1}{l|}{0.6445} &
			\multicolumn{1}{l|}{0.8891} &
			\multicolumn{1}{l|}{0.731} &
			\multicolumn{1}{l|}{0.737} &
			\multicolumn{1}{l|}{0.9403 ± 0.0081} &
			\textbf{0.9429 ± 0.0081} \\ \cline{2-8}
			\multicolumn{1}{|c|}{\multirow{-11}{*}{CIFAR-10}} &
			mean &
			%\multicolumn{1}{l|}{\textbf{0.9460}} &
			\multicolumn{1}{l|}{0.5925} &
			\multicolumn{1}{l|}{0.8718} &
			\multicolumn{1}{l|}{0.6481} &
			\multicolumn{1}{l|}{0.665} &
			\multicolumn{1}{l|}{0.9173} &
			\textbf{0.9192} \\ \hline
			
		\end{tabular}%
	}
	\vspace{1pt}
	\footnotesize{\\ 1. For Isolation Forest we use following formula: $\max(x, 1-x)$, where $x$ means AUR-ROC performance on test dataset.}
\end{table*}

\begin{table*}[t]
	\renewcommand{\arraystretch}{1.3}
	\centering
	\caption{Table with results for E2E-ConvA$^3$ and other compared methods obtained on tabular datasets.}
	\label{tab:tabular_results}
	\resizebox{\textwidth}{!}{
		\begin{tabular}{lll|l|l|l|l|l|l|l|}
			\hline
			\multicolumn{3}{|c|}{Data   Characteristics} &
			\multicolumn{7}{c|}{AUC-ROC   performance on test dataset} \\ \hline
			\multicolumn{1}{|l|}{Data} &
			\multicolumn{1}{l|}{No. of obs. \footnote{Number of observations after drop duplicates.}} &
			\% of target &
			NN classifier &
			Isolation Forest \footnote{for Isolation Forest we use following formula: $\max(x, 1-x)$, where $x$ means AUR-ROC performance on test dataset}&
			DevNet~\cite{devnet} &
			MIX~\cite{MIX} &
			A$^3$ &
			E2E-ConvA$^3$ ($\gamma = 1$) &
			E2E-ConvA$^3$ ($\gamma = 10$) \\ \hline
			\multicolumn{1}{|l|}{Credit Card Fraud Detection} &
			\multicolumn{1}{l|}{283,726} &
			0.17\% &
			0.938 &
			0.966 &
			\textbf{0.980} &
			- &
			0.975 ± 0.0162 &
			0.974 ± 0.0139  &
			\textbf{0.980 ± 0.0149} \\ \hline
			\multicolumn{1}{|l|}{kdd(http)} &
			\multicolumn{1}{l|}{580,507} &
			0.48\% &
			0.998 &
			0.944 &
			- &
			0.949 &
			\textbf{0.999 ± 9e-5} &
			\textbf{0.999 ± 0.0001} &
			\textbf{0.999 ± 5e-6} \\ \hline
			\multicolumn{1}{|l|}{kdd(ps)} &
			\multicolumn{1}{l|}{816,378} &
			0.44\% &
			0.998 &
			\textbf{0.999} &
			- &
			\textbf{0.999 } &
			\textbf{0.999 ± 6e-4} &
			\textbf{0.999 ± 3e-7} &
			\textbf{0.999 ± 6e-4} \\ \hline
			\multicolumn{1}{|l|}{CelebA} &
			\multicolumn{1}{l|}{115,115} &
			3.12\% &
			0.962 &
			0.561 &
			0.951 &
			- &
			\textbf{0.967 ± 0.0024} &
			\textbf{0.967 ± 0.0022} &
			0.966 ± 0.0014 \\ \hline
			\multicolumn{1}{|l|}{Census-Income   (KDD)} &
			\multicolumn{1}{l|}{299,285} &
			8.30\% &
			\textbf{0.926} &
			0.530 &
			0.828 &
			- &
			0.912 ± 0.0032 &  
			0.919 ± 0.0024 &
			0.918 ± 0.0033 \\ \hline
			\multicolumn{1}{|l|}{Bank   Marketing} &
			\multicolumn{1}{l|}{41,176} &
			11.27\% &
			0.937 &
			0.753 &
			0.807 &
			- &
			\textbf{0.940 ± 0.0056} &  
			0.938 ± 0.0026 &
			\textbf{0.940 ± 0.0034} \\ \hline
			\multicolumn{1}{|l|}{Thyroid   Disease} &
			\multicolumn{1}{l|}{7,129} &
			7.49\% &
			0.977 &
			0.613 &
			0.783 &
			- &
			0.993 ± 0.0054 &  
			\textbf{0.994 ± 0.0033} &
			0.993 ± 0.0047 \\ \hline
		\end{tabular}
	}
	\vspace{1pt}
	\footnotesize{\\ 2. Number of observations after deduplication, \\ 3. for Isolation Forest we use following formula: $\max(x, 1-x)$, where $x$ means AUR-ROC performance on test dataset.}
\end{table*}

\subsection{Results}

\subsubsection{Image datasets}

In experiments on image datasets we used one-vs-all procedure, which means that first we designate observations from one class as normal and treat the remaining observations as anomalies, and then fit the model in this setup. As an evaluation we measure area under the ROC curve on unseen observations. In one iteration of the experiment on a given dataset, each class is used as the normal class once. 

In the end, we conducted $10$ iterations of such an experiment. Mean and standard deviation values over every iteration of the ROC AUC are presented in Table \ref{tab:image_results}. Since the results of analogous experiments are presented in ~\cite{mkd, DSVDD, DASVDD}, we compare the results of our method the these benchmarks. In Table \ref{tab:image_results} results for MKD-AD~\cite{mkd}, DSVDD~\cite{DSVDD} and DASVDD~\cite{DASVDD} come from the papers in which these methods were introduced. The results for the Isolation Forest were calculated on-site. The best result for each dataset has been bolded.

\subsubsection{Tabular datasets}

For tabular datasets we fitted End-to-End Convolutional A$^3$ 10 times (for each value of the $\gamma$ parameter) to prevent some randomness of results. Mean and standard deviation values of the area under ROC curve are presented in Table \ref{tab:tabular_results}. The results for DevNet~\cite{devnet} and MIX~\cite{MIX} given in the table come from the papers in which these methods were introduced. The results for the other methods (NN classifier, Isolation Forest and A$^3$) were obtained by the authors. 
%For end-to-end convolutional A3, we used the following architecture: autoencoder: 1D Convolutional AE, activation type: pooled to minimal size. The best result for each dataset has been bolded.

\section{Discussion and Conclusions}

We introduce a highly non-trivial extension of a very interesting and promising approach to anomaly detection problems, combining the cutting-edge retrieval of information from the hidden layers' patterns with well known and researched AE reconstruction error methods. The two part architecture inherited from A$^3$ is augmented with the reconstruction error score, rendering it a best-of-both-worlds kind of solution, with a visible strengthening of performance.

The results of the experiments showcase that our proposed method is effective, versatile and consistent. Being essentially a feed-forward network it is lightweight and relatively uncomplicated with a straightforward learning process, yet proves itself to be more accurate and less prone to high volatility than other models with comparable complexity. Moreover, being compatible with both 1- and 2-dimensional data (with a natural extension to three dimensions also on the table), it can be easily adapted to a vast variety of real-life problems and applications, especially those where the simplicity of the design an low number of parameters is crucial. The consistency of the approach is further emphasised by the fact that it performs well in scenarios with both high and low density of anomalies.

\vspace{1cm}

* Corresponding author: p.twardowski.7@gmail.com

\end{document}